\documentclass[11pt]{article}

\usepackage[]{acl}

\usepackage{times}
\usepackage{latexsym}
\usepackage{tabularx}

\usepackage[T1]{fontenc}
\usepackage[utf8]{inputenc}

\usepackage{microtype}
\usepackage{graphicx}

\title{An Ensemble Method Based on the Combination of Transformers with Convolutional Neural Networks to Detect Artificially Generated Text}

\author{Vijini Liyanage \and Davide Buscaldi\\
  LIPN , Université Sorbonne Paris Nord, CNRS UMR 7030\\
  99 av. Jean-Baptiste Clément, 93430 Villetaneuse, France \\
  \texttt{\{liyanage,davide.buscaldi\}@lipn.univ-paris13.fr}  }%\\\And
\begin{document}
\maketitle
\begin{abstract}

Thanks to the state-of-the-art Large Language Models (LLMs), language generation has reached outstanding levels. These models are capable of generating high quality content, thus making it a challenging task to detect generated text from human-written content. Despite the advantages provided by Natural Language Generation, the inability to distinguish automatically generated text can raise ethical concerns in terms of authenticity. Consequently, it is important to design and develop methodologies to detect artificial content. In our work, we present some classification models constructed by ensembling transformer models such as Sci-BERT, DeBERTa and XLNet, with Convolutional Neural Networks (CNNs). Our experiments demonstrate that the considered ensemble architectures surpass the performance of the individual transformer models for classification. Furthermore, the proposed SciBERT-CNN ensemble model produced an F1-score of 98.36\% on the ALTA shared task 2023 data. 
\end{abstract}

\section{Introduction}

Nowadays, people have access to state-of-the-art LLMs which help them simplify some of their daily activities. One of the most notable breakthroughs in recent years is the evolution of OpenAI’s GPT models which are capable of generating text that looks as if they are written by a human. Especially, the latest models such as ChatGPT and GPT4 \cite{openai2023gpt4} have won global attention for providing solutions to any kind of question or concern that humans possess. Moreover, these models produce outputs that appear to be written by a human. 

Thus there is a potential risk in determining the authenticity of textual content that mankind refers to. Especially, in a domain such as academia, leveraging generation models in composing articles might raise an ethical concern. For example in ICML 2023, they have included a note under the “Ethics'' section prohibiting the use of text generated by
ChatGPT and other LLMs, unless “presented as part of the paper’s experiential analysis.” \footnote{\url{https://icml.cc/Conferences/2023/llm-policy}}. Accordingly, it is essential to have mechanisms for detecting artificially composed text from human written text.

Currently, a substantial amount of research has focused on the detection of automatically generated text. Recent research (\cite{zellers2019defending},  \cite{glazkova2022detecting} and  \citet{liyanage2023detecting}) mostly consider detection as a binary classification task and leverage SOTA classification models to distinguish machine-generated text from original text. Besides, some employ statistical detection tools such as GLTR \cite{gehrmann2019gltr} or latest deep learning based tools such as GPT2 output detector\footnote{\url{https://openai-openai-detector--5smxg.hf.space}}, DetectGPT \cite{mitchell2023detectgpt} or GPTZero \footnote{\url{https://gptzero.me/}}. Moreover, several researchers (\citet{liyanage2022benchmark}, \cite{kashnitsky2022overview}) have published corpora composed of machine-generated content, which can be utilized by future research on detection.

Our work is based on the participation of our team in the ALTA shared task 2023 \cite{Diego}
%\footnote{\url{https://www.alta.asn.au/events/sharedtask2023/}}. 
The objective of the task is to build automatic detection systems that can discriminate between human-authored and synthetic text generated by Large Language Models (LLMs). Their corpus is composed of artificial contents that belong to a variety of domains (such as law, medical) and are generated by models such as T5 \cite{2020t5} and GPT-X.

This paper is organized as follows. We provide the corpus and task description in Section 2. In Section 3, we describe our methodology and Section 4, deliver the experimental setup and the official results. Section 5 concludes this paper.

\section{Task Overview}

\subsection{Task Definition}

The task at hand revolves around distinguishing between automatically generated and human-written texts. In essence, it involves a binary classification challenge where the goal is to categorize provided texts into two distinct and exclusive groups. To outline this formally:
\begin{itemize}
    \item Input: We are presented with text segments.
    
    \item Output: The objective is to assign one of two possible labels to each text segment: either "human-written" or "machine-generated".
\end{itemize}

This undertaking aims to establish a clear boundary between texts created through automated processes and those crafted by human authors. The primary aim is to develop a model that can effectively differentiate between these two categories based on the characteristics of the given excerpts.

\subsection{Corpus}

The dataset published for the ALTA shared task is a balanced one composed of 9000 original (human written) excerpts and 9000 fake (artificially generated) excerpts. On average, the excerpts consist of 35 words each. To gain a deeper comprehension of the corpus, category-wise (original vs generated) statistics with respective example excerpts are provided in Table \ref{table1}.

\begin{table*}
\centering
\begin{tabularx}{\textwidth}{|c|X|X|}
\hline
\textbf{} & \textbf{Original} & \textbf{Generated}\\
\hline
Min. word count & 10 & 1 \\
\hline
%Example excerpt with min. word count & Talk to them calmly about your desire to do it. &  Reactions \\
% & & \\
%\hline
Max. word count & 96 & 192 \\
\hline
Avg. word count & 25 & 45 \\
\hline
Example excerpt & This is the data I collected so far (motorcycle standing on central stand, back wheel revolving, velocity comes from the back wheel, ABS LED blinking). & %On the other hand, the newspaper "Sabah" emphasizes that one of the perpetrators is said to have been an informant for the Federal Office for the \\%Protection of the Constitution and expresses concern that German authorities could also be involved in the right-wing extremist swamp.\\ 
In this sense, she emphasized that it was a mistake to tie development aid to times of economic booms, as it is a "permanent commitment".\\
\hline
\end{tabularx}

\caption{Statistics of the ALTA shared task corpus (The avg. figures are rounded off to the nearest whole number)}
\label{table1}
\end{table*}

\section{Methodology}

Given that the shared task frames detection as a binary classification challenge, we utilized a range of classification models to address this objective.  In the subsequent subsections, in-depth explanations are provided pertaining to the examined statistical, recurrent and transformer models, and the corresponding ensemble architectures.

\subsection{Statistical Models and their Respective Ensemble Architectures}

In our work, we primarily employed Naive Bayes, Passive Aggressive and Support Vector Machine (SVM), which are classification algorithms used in machine learning to categorize data points into different classes \cite{bishop2006pattern}. Naive Bayes is a probabilistic classification algorithm based on Bayes' theorem and it is widely used for tasks such as spam detection. It assumes that the features are conditionally independent given the class label.  Passive Aggressive is a type of algorithm that aims to make aggressive updates when it encounters a misclassified point and passive updates when the point is correctly classified. SVM is a powerful supervised machine learning algorithm used for classification and regression tasks. It is a popular algorithm in text classification tasks. These algorithms were employed in conjunction with the two text encoding methodologies, namely Bag of Words (BoW) and Term Frequency-Inverse Document Frequency (TF-IDF).

Furthermore, we harnessed the capabilities of ensembles comprising the aforementioned statistical models, applying various ensemble methodologies such as voting, stacking, bagging, and boosting. By amalgamating the predictions of multiple models, ensemble techniques aim to enhance the overall predictive power of our system. Voting combines the outputs through a majority or weighted decision, stacking involves training a meta-model on the predictions of base models, bagging leverages bootstrapped subsets of data for training individual models, and boosting iteratively adjusts model weights to prioritize difficult-to-classify instances. Through these ensemble strategies, we sought to extract richer insights from our data and attain improved classification performance.

\subsection{Recurrent Models and their Respective Ensemble Architectures}

Recurrent models, a subset of neural network architectures, are models designed to capture temporal dependencies and patterns within sequences. We conducted experiments with LSTM and Bi-LSTM models,  which are a type of RNN architecture specifically designed to address the vanishing gradient problem that can occur in traditional RNNs. To further improve classification accuracies of these models, we ensembled them with a Convolutional Neural Networks (CNNs) architecture. This approach helps in enhancing the predictive capabilities overall model by capitalizing on their respective strengths in capturing temporal dependencies and spatial features. We trained the entire ensemble end-to-end, allowing the network to learn how to best combine the features extracted by both LSTM and CNN components.

\subsection{Transformer Models and their Respective Ensemble Architectures}

For our classification experiments, we leveraged cutting-edge transformer models, namely BERT, SciBERT, DeBERTa, and XLNet. These state-of-the-art architectures have demonstrated exceptional proficiency in a wide spectrum of natural language processing tasks, including classification. BERT (Bidirectional Encoder Representations from Transformers) \cite{devlin2018bert} introduces bidirectional context by pretraining on a massive corpus and then fine-tuning on task-specific data. SciBERT \cite{beltagy2019scibert} is specialized for scientific text, adapting BERT's embeddings to domain-specific language. DeBERTa (Decoding-enhanced BERT with Disentangled Attention) \cite{he2020deberta} enhances attention mechanisms, capturing dependencies among words more effectively. XLNet \cite{yang2019xlnet} employs a permutation-based training approach to capture bidirectional context and alleviate BERT's limitations.

Initially, we created ensembles by combining the capabilities of SciBERT and DeBERTa models with the foundational BERT model. This process involves channeling the data through each base model, which comprises the transformer block along with a subsequent max pooling layer. Subsequently, the outcomes derived from these individual models are concatenated to generate a unified representation, which is then channeled into a linear classification layer for making refined predictions. 

\begin{figure*}
    \centering
    \includegraphics[width=15cm]{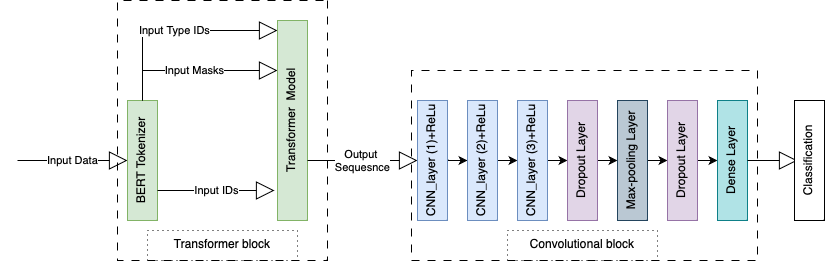}
    \caption{Architecture of Transformer-CNN Ensemble}
    \label{fig1}
\end{figure*}

Furthermore, we integrated the transformer models with Convolutional Neural Networks (CNNs) to construct ensemble architectures that exhibit enhanced performance. As depicted in the architectural diagram \ref{fig1}, the embeddings produced by the transformer model are directed into a CNN layer. In our approach, the labels won't be actively employed since we're solely utilizing the BERT transformer without a distinct specialized component on the upper layer. The CNN, placed as the upper layer, takes on the role of our main component. We exclude the nn.Embedding layers, as the need for a lookup table for embedding vectors is obviated. Instead, we directly infuse the embedding vectors from BERT into the CNN architecture.

\section{Experiments and Results}

The text underwent preliminary processing, involving the elimination of stopwords and stemming, before being supplied to either statistical or neural network architectures. The processed data was then transformed into numerical vectors using Bag of Words (BoW) or tf-idf encoding techniques, which were subsequently utilized as inputs for the statistical models. All of the employed statistical models, as well as their corresponding ensemble methods, were imported from the Scikit-learn library. %\footnote{\url{https://github.com/scikit-learn/scikit-learn}}.  
For constructing LSTM and CNN models, the relevant layers were imported from TensorFlow's Keras module. Training these recurrent models, including those combined with CNN ensembles, involved running 10 epochs. The LSTM and Bi-LSTM architectures were trained using batch sizes of 64 and 128, respectively.

Concerning transformer architectures and their associated ensembles, pre-trained models from Hugging Face \cite{wolf2020transformers} were imported and subsequently fine-tuned through the utilization of Simple Transformers \footnote{\url{https://simpletransformers.ai}}. The BERT tokenizer was consistently employed across all models. The fine-tuning process involved 3 epochs, a batch size of 16, and a maximum sequence length of 128. Leveraging the T4 GPU Hardware accelerator, the average training time for models was approximately 30 minutes. For standalone models, the input consisted of unprocessed text, while ensembles underwent pre-processing involving punctuation removal and conversion to lowercase. As represented in Figure \ref{fig1}, the CNN block of the ensembles was composed of three convolutional layers.

The dataset was split in 80:20 ratio for training and testing. 
%The experiments were conducted within a Python 3 Colab environment utilizing the T4 GPU Hardware accelerator.
To assess the classification performance of the models under consideration, the F1 score was employed. This score, being a balanced combination of precision and recall, offers a comprehensive evaluation. Each model underwent a total of five experimental iterations, and the resultant average F1 scores are presented in Table \ref{table2}.

\begin{table}
\centering
\begin{tabular}{lc}
\hline
\textbf{Model} & \textbf{F1}\\
\hline
\textbf{Statistical Models} & \\
NB + BoW & 89.04\\
PA + BoW & 84.07\\
SVM + BoW & 87.51\\
NB + tf-idf & 89.02\\
NB + tf-idf & 91.00\\
NB + tf-idf & 91.42\\
\textbf{Ensembles of Statistical Models}  \\
Voting (NB + PA + SVM) + BoW & 90.29\\
Stacking (NB + PA + SVM) + BoW &  88.23\\
Bagging (NB + PA + SVM) + BoW & 91.56\\
Boosting (NB + PA + SVM) + BoW & 90.28\\
\hline
\textbf{Recurrent Models} & \\
LSTM & 49.08\\
Bi-LSTM & 90.58\\
\textbf{Ensembles of RNNs} & \\
LSTM + CNN &  49.08\\
Bi-LSTM + CNN & 90.02\\
\hline
\textbf{Transformer Models} & \\
BERT\begin{math}_{base}\end{math} & 90.81\\
SciBERT & 94.89\\
DeBERTa\begin{math}_{large}\end{math} & 96.67\\
XLNet\begin{math}_{large}\end{math} & 93.62\\
\textbf{Ensembles of BERT models} & \\
BERT\begin{math}_{base}\end{math} + SciBERT & 97.80\\
BERT\begin{math}_{base}\end{math} + DeBERTa\begin{math}_{large}\end{math} & 97.47\\ 
\textbf{Ensembles of transformers with CNN} & \\
BERT\begin{math}_{base}\end{math} + CNN & 97.42\\
SciBERT + CNN & 97.56\\
DeBERTa\begin{math}_{large}\end{math} + CNN & \textbf{98.36}\\
XLNet\begin{math}_{base}\end{math} + CNN & 97.44\\
\hline
\end{tabular}
\caption{Classification Scores}
\label{table2}
\end{table}

In general, the ensemble architectures have exhibited superior performance compared to their corresponding original models. Our best-performing solution is the combination of DeBERTa\begin{math}_{large}\end{math} with CNN, achieving an F1 score of \textbf{98.36\%}.

\section{Conclusion}

In this work, we have explored the application of different SOTA classification models on the detection of automatically generated text from human written text. Moreover, we have created various ensemble methods with the aforementioned models and examined their performance on the detection task. Our results on the test data showed that generally the ensemble architectures outperform the considered original models. %To date, the results with our DeBERTa\begin{math}_{large}\end{math}-CNN ensemble are ranked third on the ALTA shared task.  

As future work, we plan to examine the applicability of our ensemble architectures in detecting artificially generated text in multilingual corpora. Another potential research direction involves assessing the effectiveness of knowledge-based approaches for detecting artificial text.

%\section*{Acknowledgements}

\bibliography{anthology,custom}

%\appendix

%\section{Example Appendix}
%\label{sec:appendix}

%This is an appendix.

\end{document}